# A Review of Fairness and a Practical Guide to Selecting Context-Appropriate Fairness Metrics in Machine Learning

Caleb J.S. Barr

Department of Mechanical Engineering, University of Canterbury, caleb.barr@pg.canterbury.ac.nz

Olivia Erdelyi

School of Law, University of Canterbury, olivia.erdelyi@canterbury.ac.nz

Paul D. Docherty

Department of Mechanical Engineering, University of Canterbury, Institute of Technical Medicine, Furtwangen University, paul.docherty@canterbury.ac.nz

Randolph C. Grace

Department of Psychology, University of Canterbury, randolph.grace@canterbury.ac.nz

Recent regulatory proposals for artificial intelligence emphasize fairness requirements for machine learning models. However, precisely defining the appropriate measure of fairness is challenging due to philosophical, cultural and political contexts. Biases can infiltrate machine learning models in complex ways depending on the model's context, rendering a single common metric of fairness insufficient. This ambiguity highlights the need for criteria to guide the selection of context-aware measures—an issue of increasing importance given the proliferation of ever tighter regulatory requirements. To address this, we developed a flowchart to guide the selection of contextually appropriate fairness measures. Twelve criteria were used to formulate the flowchart. This included consideration of model assessment criteria, model selection criteria, and data bias. We also review fairness literature in the context of machine learning and link it to core regulatory instruments to assist policymakers, AI developers, researchers, and other stakeholders in appropriately addressing fairness concerns and complying with relevant regulatory requirements.

## 1 INTRODUCTION

Artificial intelligence (AI) has a growing influence on our daily lives. In the past 20 years, it has been used in high-stakes decision-making, such as predicting prisoner recidivism [1], [2], evaluating creditworthiness [3], and predicting financial fraud [4]. Recently, the progression of AI and improved societal trust have enabled its use in more high-risk scenarios, such as autonomous driving [5], piloting fighter jets [6], and identifying cancer cells [7], [8]. Industrial and political demand for low-cost decision support has led to the development of novel AI approaches and implementations. However, there has also been a growing level of concern regarding the fairness of machine learning (ML) assisted decisions. For example, ProPublica's report identified racial bias in the Correctional Offender Management Profiling for Alternative Sanctions (COMPAS) software [9], a recidivism prediction tool used in parts of the United States. They found that African American offenders were twice as likely as white offenders to be labelled high risk for reoffending but not reoffend. Conversely, white offenders were more frequently labelled low risk but reoffended. Several papers refuted ProPublica's claim that COMPAS was biased against African Americans [10], [11]. Nevertheless, the ProPublica report, as well as other prominent cases of bias—like Google's ad-targeting system that showed ads for high-paying executive jobs more often to male job seekers than equivalent female job seekers [12] and Amazon's AI recruitment tool that favored male candidates [13]—instigated a proliferation of ML fairness research.

Prominent research domains include definitions of fairness in ML [14], [15], incompatibilities between fairness measures [16], [17], [18], fairness-enhancing augmentations to ML approaches [19], [20], [21], and choosing appropriate fairness measures for given contexts [22], to name a few. This quickly developing field is hard to navigate for experts and does not seem to be well-understood by most practitioners. However, the situation is currently radically changing with the emergence of increasingly onerous AI regulations worldwide. These regulations compel any stakeholder in the AI domain to learn about bias and fairness issues as part of their effort to comply with regulatory requirements, be it to keep their competitive edge, maintain their customers' trust, or avoid fines. Existing AI tools may need to be audited to confirm compliance with emerging legislation. For example, AI-assisted recruitment tools, recidivism tools, or advertisement targeting tools may yield reputational and litigation risk if they are found to be unfair across race or gender. The common element in all these settings is that the relevant teams within those organizations need to familiarize themselves with regulatory requirements on bias and fairness to carry out their respective tasks and ensure that the AI systems they develop, purchase, or use are fair.

Fundamental issues that organizations must be considered include: How are bias and fairness defined and measured? Are they the same thing, or, if not, how do they differ and what is their relationship to each other? What requirements must be observed for an AI system to be classified as fair or bias-free? Is the latter even possible? Where can these requirements be found?

Most organizations located in the EU often begin by looking at the EU AI Act (AIA) [23], only to find that it does neither define the notion of bias nor that of fairness. Carefully reading through the entire AIA, they find references to these concepts: Recital 27 mentions "diversity, non-discrimination and fairness"—one of the ethical principles of the Ethics Guidelines for Trustworthy AI developed by the High-Level Expert Group on AI (AI HLEG) [24] explaining that it means "AI systems are developed and used in a way that includes diverse actors and promotes equal access, gender equality and cultural diversity, while avoiding discriminatory impacts and unfair biases." Some provisions state requirements related to bias and/or fairness without even mentioning any of those terms. For example, Article 11 requires that technical documentation is drawn up before a high-risk AI system is put on the market or into service, with Annex IV 3 revealing more details on the contents of that documentation. Article 13 3(b)(v) aims to ensure the transparency of high-risk AI systems by requiring that they are accompanied by instructions for use. Both the technical documentation and instructions for use must contain information on an AI system's "performance regarding specific persons or groups of persons on which the system is intended to be used". Putting aside the difficulty of finding these provisions, let alone interpret them, they leave modeling teams in a predicament, as they are responsible for achieving a technical demand that lacks specificity and are given a paucity of actionable guidance.

Regulation in AI can be politically fraught. For instance, organizations in the US were guided by President Biden's Executive Order [25], however the current president has removed these obligations. There are sporadic state or local laws, usually tackling specific AI regulatory aspects. For example, the Rules of the City of New York (RCNY) [26] lays down detailed requirements on bias audits of automated employment decision tools (AEDTs). Both regulations only refer to bias, with the latter providing a reasonable level of detail on the sorts of criteria that may be relevant for assessing bias in the particular context of AEDTs. That said, neither instrument actually defines the notion of bias or lays down generally applicable criteria on bias mitigation, which could be applied throughout the entire lifecycle of an AI system and adapted to different contexts.

While frustrating for the modeling teams, the fact that high-level regulations like the AIA or the Executive Order do not provide detailed, practical implementation guidance is not surprising. In fact, it is fully in line with their objective to put a high-level, consistent, and flexible regulatory framework in place. The sort of practical guidance modeling teams are looking for should come from lower-level, more targeted regulations, such as standards. European standardization authorities are urgently publishing such standards for the AIA in compliance with the European Commission's standardization request [27]. Their US counterparts are similarly engaged. On the international level, the most authoritative standards in the AI field are those developed by the joint AI committee of the International Organization for Standardization (ISO) and the International Electrotechnical Commission (IEC)—ISO/IEC JTC 1 SC 42. We would like to draw attention to ISO/IEC TR 24027:2022 [28], which concerns bias in AI systems and AI-aided decision making and informs several related standards, partially still under development. 24027 is a technical report that provides very clear and accurate answers to most of the questions our teams are grappling with. Especially read in conjunction with other relevant standards, such as ISO/IEC 42001:2023 on AI management systems [29], it provides valuable, actionable guidance on managing bias and fairness across various AI lifecycle stages and contexts.

Due to the very broad, complex and technical nature of fairness in AI, standards such as the ISO/IEC TR 24027 cannot report all the pertinent information in a single publication, and often point the reader to recent relevant scientific literature. Hence, modeling teams need to deep-dive into that literature to find state-of-the art technical methods to comply with regulatory requirements. Another problem they will face when doing so is unraveling the many terminological inconsistencies between scientific and corresponding policy notions. This is a key barrier to both implementing regulatory requirements and leveraging scientific insights in practice.

Against this background, this paper aims to contribute to the fairness field in three main ways:

1. Build a bridge between regulation and science in the fairness space to support compliance with AI regulations: As alluded to earlier, the fairness field is suffering from terminological inconsistencies. This is mainly a consequence of the fact that several scientific communities and policymakers are contributing to shaping it. We review relevant scientific literature, provide clear definitions for core notions of interest, and link scientific literature to key regulatory sources to reduce the regulatory burden on stakeholders and enable them to benefit from scientific insights when complying with regulatory requirements.
2. Novel interpretation of the bias interaction loop: While understanding and defining different types of biases is important, in recent times, increasing attention has been devoted to analyzing the interaction between biases and the implications of these biases on AI-tool outcomes.
3. Novel, practical guidance for selecting context-appropriate fairness measures: It is also increasingly becoming clear that fairness notions—which constitute one way to identify unwanted biases in AI systems—perform very differently depending on the context in which they are used. Although a wide range of fairness measures have been defined [14], [22], [30], each method has a limited domain of appropriate implementation due to the context-specific complexities of ML models [18], [30]. This highlights the need for future research into finding appropriate fairness measures for given contexts [17], [22], [31], [32]. Building on seminal review papers that provide decision-making frameworks for this purpose [33], [34], we propose a flowchart to help stakeholders select the right fairness measure for their particular contexts. We also discuss key challenges of applying fairness measures.

Our paper is structured as follows: Section 2 introduces and defines the terms bias and fairness, as well as their relationship to each other. Section 3 provides a taxonomy of biases based on ISO/IEC TR 24027 and the complimentary scientific literature, introduces our augmentation to Mehrabi et al.'s (2021) model for bias interaction [35], and discusses techniques aimed at reducing bias entering AI systems. Section 4 systematically reviews fairness notions commonly deployed in ML. Section 5 discusses common challenges when measuring fairness. Section 6 introduces our flowchart for selecting context-appropriate fairness measures. Section 7 provides concluding remarks.

## 2 PRELIMINARIES

The lack of definitions or inconsistent usage of terms in policies, regulations and sometimes in scientific literature is adding an extra layer of complexity to understanding and complying with fairness requirements. One source of confusion is that even scientific literature tends to use bias and unfairness interchangeably [22], [36]. Therefore, we start by clearly defining the related but not interchangeable notions of bias and fairness based on ISO/IEC 22989 [37], ISO/IEC TR 24027 [28], and relevant scientific literature.

### 2.1 Bias

ISO/IEC 24027 [28] defines bias as "systematic difference in the treatment of certain objects, people, or groups in comparison to others." In scientific literature, similar definitions are used [22]. It is important to distinguish between the meaning of bias in the AI and social contexts, as well as between desired and unwanted bias.

In the AI context, certain biases are essential for the operation of AI systems, as they are necessary for ML algorithms to differentiate between particular situations and carry out their tasks, be it classification, clustering or something else. Take the example of a classification task, where the aim is to group offenders in two distinct groups, "recidivists" and "non-recidivists". No ML model can solve this task unless it is allowed to distinguish between offenders in a systematic manner. Such desired biases are different from unwanted biases, which can be introduced into AI systems in various ways and yield unfair system outcomes that negatively affect individuals or groups of people.

In the social context, the main focus of attention is less on the technical workings of AI systems than on how they impact society. Consequently, bias usually refers to AI system outcomes (i.e., how the use of an AI system may affect society rather than the actual system output) that cause injustice by unfairly discriminating certain individuals or groups of people. To avoid confusion, the standardization world discusses such settings under the umbrella of fairness instead of bias.

### 2.2 Fairness

ISO/IEC 22989 [37] and ISO/IEC TR 24027 [28] conceive of fairness as a "treatment, a behavior or an outcome that respects established facts, beliefs and norms and is not determined by favoritism or unjust discrimination." Note that neither standardization nor the scientific world offers a universal fairness definition. This is because the notion of fairness—both within the AI field and beyond—is notoriously complex, context-dependent, and differs widely due to philosophical, religious, cultural, social, historical, political, legal, and ethical factors [22]. Therefore, attempts to define a single, universally-applicable definition of fairness have been unsuccessful [38] and numerous fairness definitions exist.

### 2.3 The Relationship Between Bias and Fairness

Bias and fairness are related, in that fairness notions (also referred to as fairness measures or fairness metrics) can be used to identify unwanted biases in AI systems. Yet whether an AI system is fair by some metric depends on much more than whether unwanted bias is present in the system. Unwanted biases do not necessarily result in unfair outcomes, nor are unfair outcomes necessarily caused by unwanted biases. In general, achieving fairness—especially simultaneously satisfying more than one fairness notion—is not always possible and typically involves trade-offs. Thus, explicitly defining fairness objectives and selecting the most appropriate fairness notion(s) given a particular context in a transparent manner is of paramount importance.

## 3 BIAS IN AI SYSTEMS

ML systems can be susceptible to the same biases as humans because their development and implementation require human decisions [39], [40]. Recently, researchers have started developing standardized bias identification methods to aid regulators in assessing model fairness. Agarwal and Agarwal (2023) developed a seven-layer model to standardize bias assessment in ML, identifying where biases enter the system and outlining the role of the developers and users in minimizing them [41].

In scientific literature, ML bias is often divided into three categories [21], [35], [42], [43] based on the temporal location of bias occurrence, these include: data bias (during data collection), algorithm bias (during model development), and user interaction bias (during implementation). These biases form a closed loop [35], [42], [43] (Figure 1). Biases start at any of these sources and propagate through the AI system.

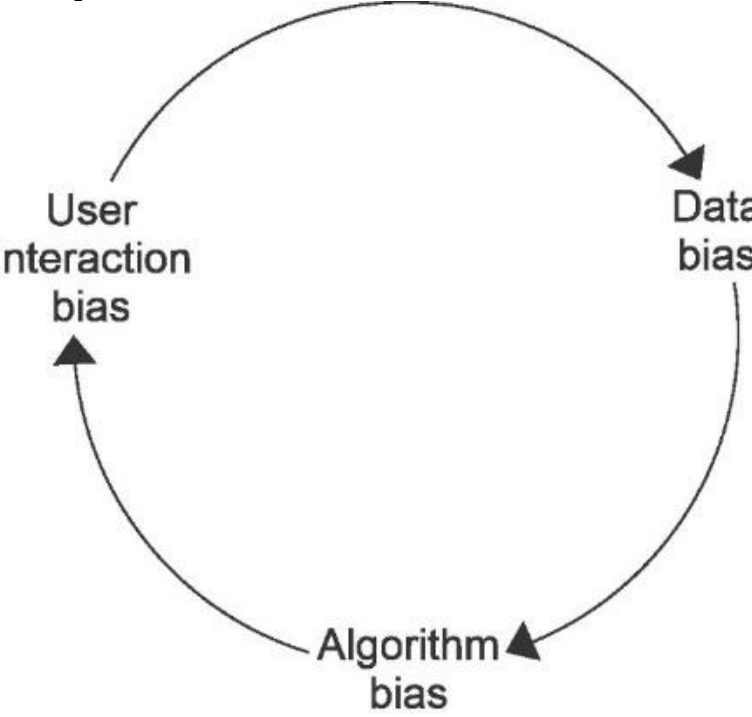

Figure 1: Bias interaction loop.

Conversely, the ISO/IEC TR 24027:2022 [28] describes bias propagation as a unidirectional interaction (Figure 2). While scientific literature defines bias temporally, the ISO/IEC TR 24027:2022 defines bias spatially (i.e., with respect to where in the AI development process these biases occur). Human cognitive biases can cause bias to be introduced by engineering decisions, or data biases.

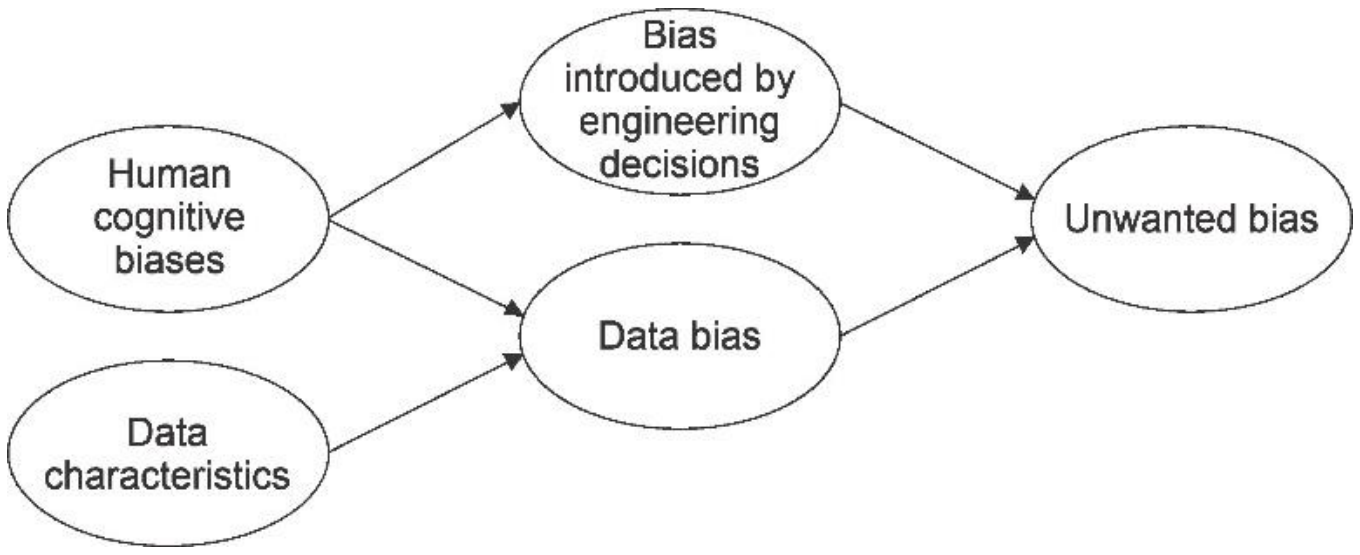

Figure 2: Relationship between high-level groups of biases as defined by the ISO/IEC TR 24027:2022 [44].

### 3.1 Data to Model Bias

Scientific literature describes data bias as biases in the dataset used to train a ML model which often lead to similar algorithm biases [35]. Among the various sources of bias, dataset bias appears to be the most frequently observed form of bias as most model decisions are decisions about data [39].

The concept of 'data bias' described in the ISO/IEC TR 24027:2022 [28] is different to that described in literature. Both scientific literature and the ISO/IEC TR 24027:2022 describe data bias as arising from technical design decisions and constraints. However, most scientific literature also considers unfair data characteristics as data biases.

#### *3.1.1Measurement Selection Bias*

Measurement selection bias occurs when predictive features are chosen and measured, leading to distortions or inaccuracies in results between protected and unprotected groups [21], [45], [46]. For example, when choosing predictive features, certain features may correlate more with protected groups (e.g., race and gender) than others [39]. This inequity can arise from disparate measurement accuracy across groups [45]. For example, on average, men typically underreport their pain in comparison to women in healthcare settings [47]. Measurement selection bias is considered a data characteristic that leads to data bias in the ISO/IEC TR 24027:2022 [28].

#### *3.1.2Omitted Variable Bias*

Omitted variable bias occurs when an important variable is excluded from the model [35], [39], [48]. Variables may be omitted because they could be proxies for protected groups or were overlooked by the modeler. For example, Mustard (2003) found that omitting conviction rates and time served from recidivism prediction models led to significant underestimation of recidivism by as much as 50% [48]. Omitted variable bias is considered a data characteristic in the ISO/IEC TR 24027:2022 [28].

#### *3.1.3Sampling and Representation Bias*

In scientific literature, sampling bias arises when the sampled population does not represent the target population [46], while representation bias occurs when the sample underrepresents part of the target population [45]. In the ISO/IEC TR 24027:2022 [28], these biases are defined under the umbrella of selection bias. An example of these sorts of biases is present in Amazon's AI recruitment tool [13], which was biased due to an underrepresentation of female applicants in the training set.

#### *3.1.4Missing Data Bias*

Martínez-Plumed et al. (2019) observed that most missing information in datasets tend to come from predictors strongly correlated with protected groups. Members of protected groups may hesitate to disclose certain information, fearing it could be used against them (described as non-response bias in the ISO/IEC TR 24027:2022 [28]). When data columns with high rates of missing data are entirely omitted from the dataset, this can lead to a lack of representation in the model for the protected group [46]. In the ISO/IEC TR 24027:2022 [28], this bias is defined as missing features and labels bias.

#### *3.1.5Aggregation Bias*

Aggregation bias (defined as data aggregation in the ISO/IEC TR 24027:2022 [28]) occurs when false conclusions about individuals are drawn from observation of group-level variables (e.g., averages and variance) [35]. False conclusions can be made when the dataset is treated under the assumption of a one-size-fits-all model when differences in populations imply they should be treated differently [45]. For example, predicting recidivism using averaged income across genders could lead to erroneous conclusions if gender-specific differences are ignored. Such as, observation at the group level might conclude that average income was positively correlated to recidivism as females have lower recidivism rates [49], [50] and lower average incomes [51] than males. In contrast, ignoring gender would most likely negatively correlate income to recidivism rate [52]. Therefore, aggregated data might suggest correlations that do not hold when considering individual-level data.

### 3.2 Algorithm to User Interaction Bias

Biases in algorithm design and construction can affect user behavior [35]. Algorithm biases fall mostly under engineering decision biases in the ISO/IEC TR 24027:2022 [28]. Engineering decision biases encompass all biases involved with model specification decisions, parameter selection and feature design.

#### *3.2.1Algorithm Bias*

Algorithmic bias (defined as model bias in the ISO/IEC TR 24027:2022 [28]) occurs when a model minimizes average error, thus fitting the model to the most typical members of a population [17], [21], [46]. Hence, the model may not be well calibrated to outlier cases. This occurs when there is sample/representation bias that is not specifically treated by the model.

#### *3.2.2Evaluation Bias*

Evaluation bias occurs during model evaluation when benchmark data used to assess a particular task does not match the target population. In such cases models are optimized on benchmark performance, and not implementation performance. Misrepresentation of benchmarks promotes models that perform well only on the benchmark data as opposed to the target population [45].

Evaluation bias falls under the algorithm bias framework of scientific literature because of its influence at model development, the ISO/IEC TR 24027:2022 [28] defines evaluation bias (termed coverage bias) as a branch of data biases because of its misrepresentation of data.

#### *3.2.3Popularity Bias*

The more an item is exposed, the more popular it becomes due to increased visibility [53], [54]. This leads to algorithm bias in models such as recommender systems. In recommender systems, popular items are more likely to be recommended to the public, which in turn increases their popularity, making them more likely to be recommended again.

#### *3.2.4Decision Bias*

Decision bias occurs during the selection of design metrics while building a ML model [39]. This includes biases introduced via the selection of the ML algorithm (termed algorithmic selection bias by the ISO/IEC TR 24027:2022 [28]), and hyperparameter tuning.

### 3.3 User Interaction to Data Bias

Many data sources for model training are generated by the developer through either feature engineering, data manipulation (such as translation and rotation of images for image recognition models), or data selection. Evidence shows that AI bias can reinforce human cognitive biases [55]. Therefore, biases generated by user interaction with model outcomes can lead to cognitive manipulation during data development, allowing the formation of data biases. User interaction biases fall under human cognitive biases in the ISO/IEC TR 24027:2022 [28].

#### *3.3.1Historical Bias*

Historical bias (labelled societal bias in the ISO/IEC TR 24027:2022 [28]) occurs when historically biased human decisions are used to generate data [17], [21], [45], [46]. For instance, historically disadvantaged groups can be arrested at higher rates [56]. In recidivism prediction models, this could lead to an overestimation of offending risk for the disadvantaged population [39], and therefore increased surveillance and further reconviction of the disadvantaged group. Thereby propagating historical biases.

#### *3.3.2Temporal Bias*

Temporal biases arise from variations in behavior over time [57]. These changes in behavior may be due to policy changes, cultural shifts, or changes in reporting. These variations can cause differences between training and implementation data, negatively affecting model performance. For example, recidivism rates noticeably changed after implementation of Washington State's Offender Accountability Act [58], which introduced a new recidivism prediction tool. The prediction tool was trained on retrospective data that was not highly indicative of future recidivism.

#### *3.3.3Population Bias*

Population bias (termed group attribution bias in the ISO/IEC TR 24027:2022 [28]) occurs when the sampled demographic does not represent the entire population's characteristics [57] due to the assumption that what is true for a small subset is true for the population [28]. This can lead to sampling and representation bias in the datasets.

#### *3.3.4Confirmation Bias*

Confirmation bias is the unconscious promotion of data, processes or model output interpretation that confirms the researcher's preconceptions [46]. This leads to outcomes that are biased towards the modeler's preconceptions, rather than capture the actual reality. This is a bias in human cognition.

### 3.4 The Impacts of Context in Bias Identification

Each AI development has its own specific model requirements and faces different framings of the AI problem. Model requirements include the specific input parameters, classification or scoring goals, and definitions of accuracy or fairness. Model framing includes consideration of the consequences of model use; how the model is used; and where the training data comes from. These considerations form the context of the approach. The outcomes of the AI model can be sensitive to that context. Thus, careful consideration of the context of an AI implementation can help with the identification of biases in an AI model.

For example, Obermeyer et al. (2019) found that a US healthcare algorithm that evaluated patient sickness for a given risk score predicted black patients to be equally as ill as white patients, despite being considerably sicker. This bias arose because the algorithm predicted healthcare costs rather than illness, assuming healthcare cost to be a proxy for illness [59]. Since there was lower access to care for black patients, their healthcare costs were lower than white patients. Hence, the framing of the scenario in terms of health costs led to inappropriate bias. Several authors have discussed the relationship between model context and bias identification [39], [45], but this has yet to be linked to the bias interaction loop. Below, we describe examples of how context interacts at each stage of the bias interaction loop.

Data to model bias can occur when the declared goal of an AI approach cannot be effectively determined by the data. For instance, using data-driven approaches to predict recidivism cannot accurately capture actual recidivism rates, but rather reports, arrests, or conviction rates [39], [60]. This is a significant issue for offences with low reporting levels [61].

Using reported recidivism introduces a natural bias, as different groups are arrested at different rates [39], [60], [62]. Hence, while framing recidivism as a simply observable quantity rather than attempting to capture the true value is convenient, and at times necessary, this can lead to bias that could be mitigated if the true incidence could be estimated or measured.

Model to user interaction bias can occur when users misinterpret or misuse the predictions of a model. For example, a user may incorrectly infer that a positive model prediction implies a higher chance of positive outcomes than indicated by the prediction threshold. In hiring processes, a manager may overvalue the scoring of a CV evaluation tool, which typically identifies desired phrases, skills, or experiences [63]. However, these tools may lack the specificity of a careful human reader, with evidence suggesting AI hiring tools may be rejecting some of the best job applicants [64]. Therefore, overreliance on AI models can introduce biases in decision making. The imperative is for the users of AI to be aware of model limitations.

User interaction to data bias can occur when a modeled outcome leads to certain biases that are then used to retrain the model [17], [22], [65]. For example, model collapse can occur when large language models (LLMs) are trained on model-generated data [66], [67]. This is particularly pertinent given the large amount of AI created content that is being passed off as original work [68]. Selbst et al. (2019) described the bias, generated by user interaction with models, as the 'ripple effect trap', where designers fail to identify how people and organizations will respond to the model [69].

These examples illustrate how specific contextual implications of an AI problem can create unique biases at each stage of the bias interaction loop. To address this, we propose a new bias interaction loop that incorporates model context into the identification of biases (Figure 3). Moreover, because bias enters a system in various ways, assessing fairness in an AI model cannot follow a 'one-size-fits-all' approach. Context is required to select the most appropriate measure of fairness that considers the specific biases occurring in the model.

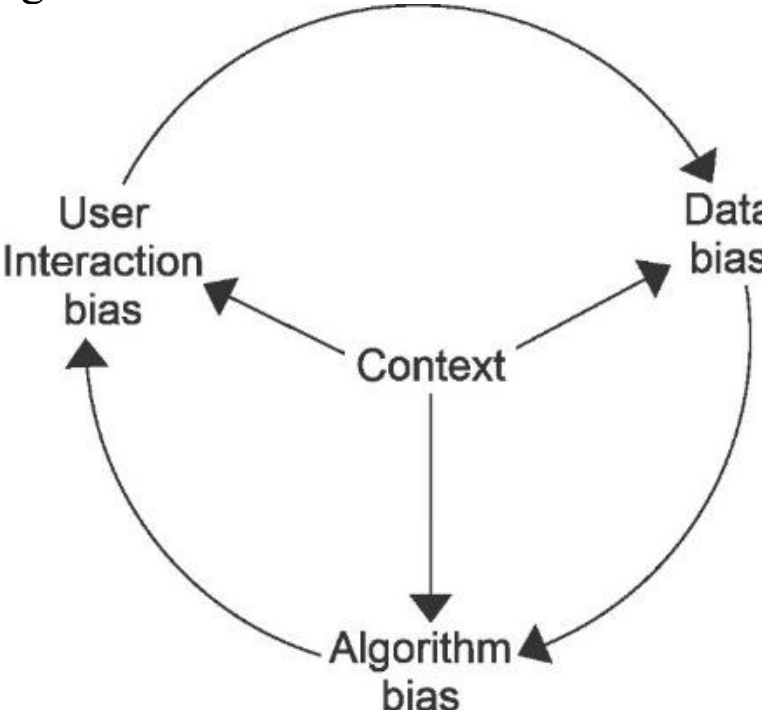

Figure 3: Bias interaction loop including model context. Specific context implications for each AI model can cause unique biases at each stage of the bias interaction loop.

### 3.5 Efforts to Reduce Bias Entering an AI System

While numerous fairness-enhancing methods have been proposed [21], [70], these methods rarely consider the causes of biases. Below we summarize the various studies on bias mitigation in AI models. Pessach and Shmueli (2022) [21], and Caton and Haas (2023) [70] provide comprehensive overviews of many of the fairness-enhancing methods.

To address incomplete datasets, Martínez-Plumed et al. (2019) suggested that mean imputation for quantitative datasets and mode for qualitative datasets would reduce bias. They argued that imputation is preferable to deletion, as rates of incompleteness in datasets can be correlated with protected groups (as explained in Section 3.1.4). The authors noted that mechanisms to handle missing data were not incorporated in many fairness-enhancing methods [46], despite consistent reports that missing data can lead to bias [21], [71], [72].

Jain et al. (2019) used singular race models to reduce algorithmic bias and improve accuracy in recidivism prediction. However, this approach resulted in reductions in fairness for certain definitions due to base rate differences between compared groups. The authors proposed future work with singular race models on balanced datasets to improve fairness [73].

Gichoya et al. (2023) provided broad principles for creating unbiased datasets in the AI medical imaging field. These principles included avoiding the exclusion of diverse and underrepresented populations, avoiding the classification of certain underrepresented populations into broad "other" categories, and avoiding the use of geographically narrow datasets [74].

Rana et al. (2023) reviewed various approaches designed to mitigate biases across the ML development cycle [43], highlighting the importance of addressing bias at every stage of ML model development.

## 4 DEFINING FAIRNESS IN MACHINE LEARNING

Definitions of 'fair' differ widely due to philosophical, religious, cultural, social, historical, political, legal, and ethical factors [22]. These factors, along with personal history, can significantly influence how a system architect incorporates fairness into a ML approach [35]. The wide range of influences contributes to the numerous fairness definitions proposed. Mehrabi et al. (2021) highlighted that most fairness definitions (in addition to datasets and problems) proposed in the literature have been developed by Western researchers [35], leading to a relatively Euro-centric perspective in fairness research.

Various software toolkits have been specifically designed to aid in the assessment of fairness in ML models: AIF360 [75]; FairLearn [76]; TensorFlow Responsible AI [77]; Aequitas [78]; and Themis-ML [79]. Among these, Aequitas is the most cited toolkit in the literature [22]. It allows users to test models for several bias and fairness metrics [78]. AIF360 provides users with bias-mitigation algorithms and offers guidance for selecting the most appropriate tool [75]. FairLearn, TensorFlow Responsible AI, and Themis-ML also provide these functionalities to varying degrees.

### 4.1 Legal Notions of Fairness

The recent emergence of fairness research has largely been driven by studies from the USA. The US legal framework is governed by two notions of fairness: disparate treatment and disparate impact (Civil Rights Act 1964). Disparate treatment refers to the differential treatment of individuals based on their group membership [80]. Disparate impact occurs when a neutral policy negatively affects members of one group more than another [80]. Satisfying both of these notions simultaneously is not always possible. For example, considering gender in a ML model would be classified as disparate treatment, but not considering gender could lead to disparate impact if there are gender differences in the predicted outcome.

Protected groups defined under US law (as described by Hardt (2020) [81]) include race [80], color [80], sex [80], [82], religion [80], national origin [80], citizenship [83], age [84], pregnancy [85], familial status [86], disability status [87], [88], veteran status [89], [90], and genetic information [91].

Other jurisdictions have different laws. For example, legal notions of fairness in NZ are defined under the Human Rights Act 1993, which prevents unfair treatment on the basis of "irrelevant personal characteristics" [92]. These personal characteristics include sex, marital status, religious belief, color, race, ethnic or national origins, disability, age, political opinion, employment status, family status, and sexual orientation [92]. The Human Rights Act 1993 ultimately requires ML models to avoid causing inferior outcomes for these protected groups.

### 4.2 Academic Notions of Fairness

To identify the most popular fairness definitions currently in use, a systematic review was conducted (Figure 4). Fairness notions implemented in each of the examined research articles were recorded, and the results presented in Table 1.

#### *4.2.1Search Strategy*

Google scholar, Scopus and PubMed databases were searched for relevant articles. Article titles were searched for the following terms: ("machine learning" OR "artificial intelligence") AND ("predict" OR "assess" OR "model" OR "predicting" OR " assessing" OR "modeling") AND ("fairness" OR "bias" OR “fair”). The search was limited to articles published between January 1st, 2023, and March 3rd, 2024. This search yielded 47 documents from Google scholar, 53 from SCOPUS, and 10 from PubMed. After removing duplicates, 76 unique publications remained.

#### *4.2.2Eligibility Criteria*

The inclusion criteria required the publication to be a journal article or full conference paper written in English. The abstracts were reviewed to ensure they indicated that measurements of fairness against a ML model were conducted. The study was ultimately included if fairness of a ML model was measured and presented. In total, 21 documents were found that met our criteria. Figure 4 presents a flowchart of the study identification results.

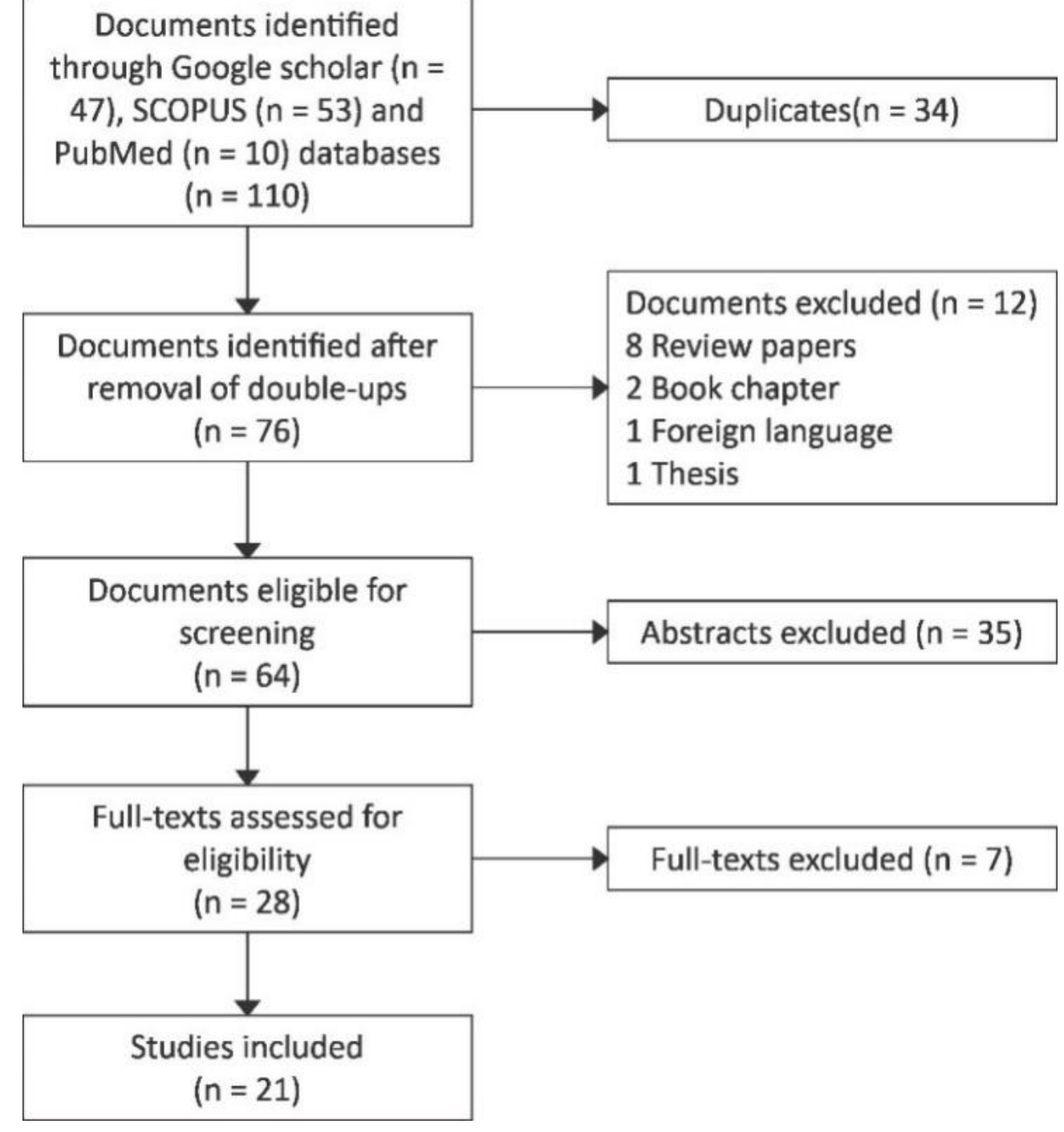

Figure 4: Flowchart of study identification results.

#### *4.2.3Fairness Notions*

Historically, fairness in ML has been primarily defined through observational measures [16], [19]. Recently, literature has proposed causal-based approaches to measuring fairness [93], [94]. Observational measures of fairness consider only the data, assessing statistical relationships between variables. In contrast, causal approaches consider a broader context, including how data is generated and how variable selection affects model behavior [33], [95], [96]. Causal fairness notions typically consider either interventions that simulate randomized experiments, or hypothetical idealized worlds (counterfactuals) that are compared to the actual world [33]. Some researchers argue that causal measures of fairness are necessary to address the problem of fairness comprehensively [33], [97], [98].

A seminal example illustrating the importance of causal models is the Berkeley admissions study [21], [97], [99]. In this study, 44.3% of men and 34.6% of women who applied to Berkeley were successful. Observational measures suggested a bias in admissions against female applicants. However, further analysis revealed that there were no significant differences in acceptance rates between males and females within individual programs. Female applicants tended to apply for courses with lower acceptance rates, accounting for global differences. The causal relationship between the global acceptance rates and gender would not have been captured by observational fairness measures. Notably, framing gender equality solely in terms of acceptance rates within programs fails to address the inequality of access to preferred programs across genders. This example underscores the complexity of defining contextual fairness in ML, highlighting the challenge of selecting appropriate fairness metrics.

Recently emerging definitions of fairness are causal-based [33], [100]. However, opponents of causal metrics argue that obtaining the correct causal model is challenging [21], with even slight changes in causal models significantly affecting fairness outcomes [101]. Furthermore, in practice, measuring causal-based fairness is difficult because measures are not always solely data-driven [33]. This is evidenced by the survey of currently employed measures of fairness (Table 1), which noted the absence of causal measures. Makhlouf et al. (2022) ranked various causal-based fairness notions on their difficulty of being deployed in practice [33]. In a series of examples, Nilforoshan et al. (2022) provided a commentary on the difficulties of applying causal-based fairness notions [102].

Observational and causal fairness measures can both be further categorized into individual or group definitions [35], [103]. Group fairness measures can also be subdivided into those with binary or regression output. Group definitions are based on comparisons between groups, whereas individual definitions are based on similarity criteria between datapoints. Some researchers evaluate fairness using both group and individual measures to provide a comprehensive view of fairness [104]. However, individual measures of fairness tend to be underrepresented in the ML fairness literature, as evidenced by Table 1, where no individual measures were reported. Table 1 categorizes each fairness metric into these six categories (observational/causal, group/individual, binary/regressive). While fairness is often categorized this way, fairness in text-to-image evaluation [105] and other generative models does not always fit these categories. Furthermore, the survey only noted instantaneous measures of fairness. Time dependent (dynamic) fairness may be more appropriate when temporal correlations to predictive outputs exist, although measuring this data is more costly. Tang et al. (2023) provide definitions of fairness in the dynamic setting [106].

$EOP$, $SP$ and $EO$ were the most common measures of fairness. This aligns with the findings of Pagano et al. (2023) [22]. These measures are easy to calculate, as they are based directly off the confusion matrix. However, the effectiveness of these approaches is controversial in cases with mismeasured base rates [60], [107], [108].

Table 1: Fairness notions implemented in recent literature as found by the systematic review.

| Metric | Num. Cit.[a] | O/C[b] | G/I[c] | B/R[d] | References |
|---|---|---|---|---|---|
| Equal Opportunity ($EOP$) | 11 | O | G | B | [109], [110], [111], [112], [113], [114], [115], [116], [117], [118], [119] |
| Statistical parity ($SP$) | 10 | O | G | B | [109], [111], [112], [113], [114], [117], [118], [119], [120], [121] |
| Equalized odds ($EO$) | 8 | O | G | B | [110], [118], [119], [122], [123], [124], [125], [126] |
| Predictive parity ($PP$) | 3 | O | G | B | [109], [110], [114], [124] |
| Balanced group balanced accuracy ($BG-BACC$) | 3 | O | G | B | [111], [115], [124] |
| Balanced group accuracy ($BG-ACC$) | 2 | O | G | B | [109], [110] |
| Equal mis-opportunity ($EMO$) | 2 | O | G | B | [109], [110] |
| Average odds ($AO$) | 1 | O | G | B | [112] |
| Balanced group $F1$ ($BG-F1$) | 1 | O | G | B | [110] |
| Balanced group $AUC$ ($BG-AUC$) | 2 | O | G | R | [110], [127] |
| Calibration ($CAL$) | 2 | O | G | R | [128], [129] |

[a] Num. Cit. = Number of Citations. [b] O = Observational; C = Causal. [c] I = Individual; G = Group. [d] B = Binary; R = Regression.

#### *4.2.3.1 Mathematical Definitions of Group Fairness with Binary Outputs*

Group fairness measures are calculated either by the absolute difference between groups of the corresponding metric or by their disparity ratios. We present all group fairness definitions as disparity ratios.

Equal opportunity ($EOP$) is a comparison of the true positive rate ($TPR$) between groups that belong to either group $x \in \{1, 2\}$:

$$TPR_n = {TP_n}/{(TP_n + FN_n)} \tag{1}$$

$$EOP = \frac{TPR_1}{TPR_2} \tag{2}$$

Statistical parity ($SP$), also known as group parity or disparate impact, is a comparison of the ratio of positive predictions ($PosPred$):

$$PosPred_n = \frac{TP_n + FP_n}{TP_n + FP_n + TN_n + FN_n} \tag{3}$$

$$SP = \frac{PosPred_1}{PosPred_2} \tag{4}$$

Equalized odds ($EO$) measures misclassification, comparing the false positive rate ($FPR$) and false negative rate ($FNR$):

$$\{FPR_n, FNR_n\} = \left\{\frac{FP_n}{TN_n + FP_n}, \frac{FN_n}{TP_n + FN_n}\right\} \tag{5}$$

$$EO = \left\{\frac{FPR_1}{FPR_2}, \frac{FNR_1}{FNR_2}\right\} \tag{6}$$

Predictive parity ($PP$) compares the positive predictive value ($PPV$)—the likelihood of a positive prediction being correct—and the negative predictive value ($NPV$) [130]:

$$\{PPV_n, NPV_n\} = \left\{\frac{TP_n}{TP_n + FP_n}, \frac{TN_n}{TN_n + FN_n}\right\} \tag{7}$$

$$PP = \left\{\frac{PPV_1}{PPV_2}, \frac{NPV_1}{NPV_2}\right\} \tag{8}$$

Balanced group balanced accuracy ($BG\text{-}BACC$) is a comparison of balanced accuracy ($BACC$)—the average of recall and specificity:

$$Recall_n = TPR_n \tag{9}$$

$$Specificity_n = {}^{TN_n}/_{(TN_n + FP_n)} \tag{10}$$

$$BACC_n = {}^1/_2(Recall_n + Specificity_n) \tag{11}$$

$$BG\text{-}BACC = \frac{BACC_1}{BACC_2} \tag{12}$$

Balanced group accuracy ($BG\text{-}ACC$) is a comparison of accuracy ($ACC$):

$$ACC_n = \frac{TN_n + TP_n}{TN_n + TP_n + FN_n + FP_n} \tag{13}$$

$$BG\text{-}ACC = \frac{ACC_1}{ACC_2} \tag{14}$$

Equal mis-opportunity ($EMO$), also called predictive equality, compares the $FPR$:

$$EMO = \frac{FPR_1}{FPR_2} \tag{15}$$

Average odds ($AO$) compares the average of the $FPR$ and $TPR$ ($ao$):

$$ao = {}^1/_2(FPR_n + TPR_n) \tag{16}$$

$$AO = \frac{ao_1}{ao_2} \tag{17}$$

Balanced group $F1$ ($BG\text{-}F1$) compares the $F1$-score—the harmonic mean of recall and precision:

$$Precision_n = PPV_n \tag{18}$$

$$F1_n = \frac{2(Recall_n \times Precision_n)}{Recall_n + Precision_n} \tag{19}$$

$$BG\text{-}F1 = \frac{F1_1}{F1_2} \tag{20}$$

#### *4.2.3.2 Mathematical Definitions of Group Fairness with Regressive Outputs*

Balanced group $AUC$ ($BG\text{-}AUC$) compares the area under the receiver operating characteristic curve between groups.

$$AUC_n = \int TPR_n \, d(FPR_n) \quad (21)$$

$$BG\text{-}AUC = \frac{AUC_1}{AUC_2} \quad (22)$$

Calibration ($CAL$) is satisfied when, for each possible risk score, the probability of a positive outcome is equal for all groups and is equal to the risk score. For example, calibration would be achieved in a recidivism prediction model if, for all groups, $h$% of people receiving a score of $h$% recidivate.

$$CAL = \frac{1}{M} \Sigma_{i=2}^{M} \left| \hat{Y}_{i-1..i} - \overline{Y}\left(\hat{Y}_{i-1..i}\right) \right| \quad (23)$$

Where the domain of prediction is discretized into $M$ bins, $\hat{Y}$ is the prediction for an individual, $\hat{Y}_{i..i-1}$ is the mean prediction of a certain bin, and $\overline{Y}\left(\hat{Y}_{i-1..i}\right)$ is the average true outcome for cases within the $i^{th}$ bin.

Balance for the positive and negative class did not appear in the survey of recent literature. However, these measures are important in situations where base rates are unequal (for more information, see Section 6.1.8). Balance for the positive ($bal-pos$) and negative ($bal-neg$) class requires the average predicted probability for people who belong to the positive and negative class to be uniform across groups. Overall balance ($BAL$) requires satisfaction of both balance for the positive and negative class.

$$\begin{Bmatrix} bal\text{-}pos_n, \\ bal\text{-}neg_n \end{Bmatrix} = \begin{Bmatrix} \frac{1}{N_{Y_n=1}} \Sigma\left(\hat{Y}_n \middle| Y_n=1\right), \\ \frac{1}{N_{Y_n=0}} \Sigma\left(\hat{Y}_n \middle| Y_n=0\right) \end{Bmatrix} \quad (24)$$

$$BAL = \left\{ \frac{bal\text{-}pos_1}{bal\text{-}pos_2}, \frac{bal\text{-}neg_1}{bal\text{-}neg_2} \right\} \quad (25)$$

Where $Y$ is the true outcome for an individual and $N$ is the number of cases where the outcome $Y = 0$ or $Y = 1$ occurred.

### 4.2.3.3 Mathematical Definitions of Individual Fairness

Individual measures of fairness did not appear in the survey of recent literature. However, these measures are historical and should be considered. Individual measures of fairness stem from Dwork et al.'s (2012) work describing fairness by the idea that "similar individuals are treated similarly" [14].

Fairness through awareness ($FTA$) is defined by k-Nearest Neighbors consistency ($KNNC$), which measures the similarity of prediction for individuals with similar attributes. This requires the attributes to be continuous so a distance metric can be used to observe similarity. Mathematical definitions were first defined by Dwork et al. (2012) [14]. However, the function defined by Zemel et al.'s (2013) [131] has become most popular.

$$FTA = KNNC = 1 - \frac{1}{N} \sum_{i=1}^{N} \left| Y_i - \frac{1}{k} \sum_{j \in N_{kNN}(x_i)} Y_j \right| \quad (26)$$

Where $N$ is the number of datapoints ($x$), $k$ is the number of neighbors considered, $Y_i$ is the outcome for a certain datapoint, and $Y_j$ the outcomes of the datapoints closest to $x_i$. A score of 1 means that it is individually fair, and a score of 0 means that it is not individually fair.

# 5 CHALLENGES WITH MEASURING FAIRNESS

Assessment of fairness in an AI model is made difficult by several factors. Firstly, to assess group-level fairness, groups that may be unfairly treated must be identified. However, unfairly treated groups may be represented by the intersections of commonly recognized marginalized groups or entirely new groups altogether. These can be hard to identify. Secondly, the application of fairness metrics in certain contexts is constrained by inherent limitations, such as base rate differences and fairness metric incompatibilities. Unequal base rates can skew the results of fairness assessments for specific metrics, undermining their validity. Furthermore, particular characteristics in data across groups can make it impossible to satisfy two fairness notions simultaneously, while avoiding trivial solutions [16], [17], [18], [22]. Finally, the optimization of models for improved fairness can lead to a reduction in model accuracy due to the fairness-accuracy trade-off.

## 5.1 Problems with Marginalized Group Identification

In a review of bias identification studies in AI, Bucchi and Fonseca (2023) found that most studies identifying biases in AI used an a priori approach. This method identifies biases in marginalized groups where known social biases exist [132]. Studies employing an a priori approach (see [40], [105]) can effectively identify ML model biases among these recognized marginalized groups. However, this preconceived notion of biases targeting predefined groups limits the scope of bias identification. Bucchi and Fonseca (2023) identified only one study that used an a posteriori approach to identifying biases (i.e., without assuming predefined marginalized groups [132]). Using this method, Watkins (2023) discovered a previously unidentified marginalized group (an unguided cluster of subjects later described as 'workers') [133]. Such discovery would have been impossible using an a priori approach. Therefore, AI researchers should be careful not to limit their perspective on what groups may be marginalized in certain AI implementations. Selbst et al. (2019) described this problem as the 'solutionism' trap that designers and researchers fall into when examining fair AI. This is the failure to identify new problems and new solutions, or in this case new groupings of marginalization, within an AI task. It is akin to the "law of the instrument" [69] where "if you have a hammer, everything looks like a nail" [134].

## 5.2 Base Rates

A group's base rate is their percentage of positive outcomes. Fairness measures defined by the confusion matrix require base rates to be uniform across groups [18] due to the relationship between outcome predictions and thresholding. Grant (2023) provides an example illustrating the effect base rates have on fairness metrics in a recidivism prediction scenario. In this example, professional thieves had a 90% chance of reoffending compared to 10% for amateur thieves. Suppose Group 1 contained 100 professionals and 10 amateurs, and Group 2 contained 10 professionals and 100 amateurs. A straightforward method of predicting recidivism would be to assign a high-risk score to professionals and a low-risk score to amateurs. This would result in Group 1 (base rate = 0.83) having a $FPR$ of 53% and a $FNR$ of 1%, while Group 2 (base rate = 0.17) would have a $FPR$ of 1% and a $FNR$ of 53% [107]. To achieve fairness across Group 1 and Group 2 using $EO$ (6), the $FPR$ and $FNR$ rates should be equal across groups. This is not possible while maintaining any level of model accuracy. This shows that even with a valid prediction system, fairness measures defined by the confusion matrix are skewed in the direction of the base rate (e.g., lower base rates lead to more false negatives). Therefore, under unequal base rates, it is impossible to determine whether unfairness is due to biased data or the base rate itself. Many authors claim equal base rates are a requirement for fairness measures defined by the confusion matrix [60], [108], [135]. As a result, the use of certain fairness metrics is limited when there is disparity in group base rates.

### 5.3 Incompatibility Results

The disparity in base rates across groups can lead to incompatibilities between certain fairness metrics. This was the focus of much of the earlier work on fair ML. Chouldechova (2017) mathematically proved that $PPV$ is incompatible with $EO$ when compared groups have unequal base rates [60]. Barocas et al. (2019) [96] and Wasserman (2004) [136] further demonstrated the incompatibility between $NPV$ and $EO$ unless a model is perfectly predictive. Kleinberg et al. (2016) and Pleiss et al. (2017) showed that $CAL$ cannot be simultaneously satisfied with $EO$ unless predictions are perfect or base rates are equal [108], [137]. $SP$ and $BG - ACC$ were found to be incompatible when base rates are unequal [30]. $SP$ and $EO$ are also incompatible unless predictions are perfect [30]. Finally, Garg et al. (2020) mathematically proved that $SP$, $EO$, and $PP$ are jointly incompatible unless base rates are uniform [138]. Furthermore, it is worth noting that equivalences exist between certain fairness metrics. For example, when $EO$ criteria are met, the $TPR$ is equal between groups, satisfying $EOP$ requirements. A detailed investigation of these incompatibilities across various metrics is presented in [30].

#### *5.3.1Empirical Example of Incompatibilities*

ProPublica demonstrated the inequality of COMPAS predictions using the $EO$ framework [9]. However, several articles argued that $EO$ was an inappropriate measure of fairness due to unequal base rates, suggesting that $CAL$ [11], [139], [140], $BG - AUC$ [10], [11], or $PP$ [10] should be used instead. Evaluation of these metrics on the same model did not reveal any unfairness in COMPAS predictions, highlighting that these measures cannot be simultaneously satisfied with $EO$. Consequently, it is now widely accepted that using $EO$ to assess fairness is inappropriate for the contextual scenario presented by COMPAS. Grant (2023) and Green (2022) provide detailed explanations of the incompatibilities between fairness metrics in the context of COMPAS [107], [141].

#### *5.3.2Consequences of Incompatibilities*

To address the consequences of incompatibilities, several authors have proposed methods to balance the trade-offs between fairness measures [141]. One approach is the relaxation of fairness requirements [34], [137]. For instance, Pleiss et al. (2017) relaxed the criteria for both $EO$ and $CAL$ to allow approximate satisfaction of both metrics simultaneously [137]. However, this partial satisfaction of incompatible fairness metrics resulted in significantly lower accuracy solutions [137]. Kleinberg et al. (2018) suggested that observing all fairness measures individually and carefully analyzing their effects would enable precise quantification of the trade-off and their acceptability within the local legal framework and culture. They argue that this approach would lead to more purposeful judgements about the appropriate use of AI in society [141], [142]. However, this is an arduous task, and one that would require a high level of expertise and investigation for every AI implementation. Hence, it may be more practical to define the types of metrics that should be considered in certain general fields.

These consequences have significant implications for measuring fairness in ML. In the academic space, several authors have expressed skepticism about whether achieving a 'fair' model is possible [16], [108], [143], [144], [145]. While some authors view these incompatibilities as a chance for society to innovate and consider fairness more closely when developing models [142]. While in the regulatory space, incompatibilities mean that most ML models can be classed as both fair and unfair depending on the definition considered. This could lead to disputes between modelers and regulators, as both could feasibly be correct in arguing whether a model is fair or not. For that thought, some argue that decisions regarding fairness should be left to policymakers [16], [146]. Therefore, it is crucial for the academic community to provide clear guidance on which fairness metrics are appropriate for various AI applications. The trade-offs between fairness, and the societal and ethical consequences of using each measure must be clearly defined. Careful consideration of the context is also critical.

This necessity has led to the development of different methods for selecting context-appropriate fairness measures [33], [34].

### 5.4 The Fairness-Accuracy Trade-off

Fairness in ML is often considered alongside accuracy due to the widely observed fairness-accuracy trade-off [16], [19], [139], [147], [148], [149], [150], [151], [152]. For ML models with optimized accuracy, increasing fairness necessitates a degradation in accuracy. Excluding trivial cases, it is impossible to maximize both fairness and accuracy simultaneously [16]. This is because fairness requirements act as additional constraints when training a model [21]. Menon and Williamson (2018) note that the magnitude of the fairness-accuracy trade-off is proportional to the correlation between the sensitive attribute (the group classifier, e.g. gender) and the target variable [150].

Conversely, recent research has challenged the assumption that a fairness-accuracy trade-off always exists. For instance, Langenberg et al. (2023) demonstrated that for balanced datasets (i.e., containing equal numbers of positive and negative outcomes), increasing accuracy improved fairness [153] due to the minimization of errors [154]. There are other instances where the fairness-accuracy trade-off is negligible [155], [156]. Despite these isolated cases, a trade-off between fairness and accuracy is generally accepted as ubiquitous in ML [16], [150], with the strength of the trade-off dependent on the context of application. Hence, the level of influence of the fairness metric on model accuracy should be carefully controlled. Typical measures of model accuracy include $AUC$ [157], $F1$-score [158], [159], and accuracy ($ACC$) [160], [161].

## 6 SELECTING APPROPRIATE FAIRNESS MEASURES BASED ON CONTEXT

The importance of defining context-appropriate fairness metrics is presented in [21], [31], and [162]. The impacts of different fairness requirements on accuracy are inconsistent [16], highlighting the need for appropriate fairness measures to balance this trade-off. Furthermore, unequal base rates skew certain fairness measures, making it impossible to determine whether unfairness was the result of biased data or the base rate itself. This limits the set of context-appropriate fairness metrics under unequal base rate conditions. Lastly, the incompatibility between fairness measures can lead to contradictions. Hence, a framework for selecting the most appropriate fairness metric for the given context is required.

Makhlouf et al. (2021 & 2022) developed novel methods for selecting context-appropriate observational [34] and causal [33] fairness notions using carefully constructed flowcharts. However, fairness in ML is a rapidly evolving research field and the outcomes of the papers reviewed suggest that different decision metrics may be appropriate. These include consideration of the type of fairness assessment [106], type of model [163], type of output [65], base rate [18], and data biases (Section 3). Figure 5 provides a flowchart for selecting context-appropriate observational fairness notions based on the reviewed contextual fairness literature.

### 6.1 A Context-Based Method for Selecting Observational Fairness

Twelve criteria were used to formulate the flowchart in Figure 5. These criteria include consideration of the model assessment criteria, such as whether fairness of data or fairness of outcome is being assessed (Section 6.1.1); whether equity requirements are in place (Section 6.1.5); whether thresholds are fixed or floating (Section 6.1.7); whether there is an emphasis on precision or recall (Section 6.1.9); whether there is an emphasis on FP or FN misclassification (Section 6.1.10); whether there is an emphasis on the positive or negative class (Section 6.1.11); and whether the dataset is balanced (Section 6.1.12). The selection criteria also include consideration of model design, such as whether the model is used for classification tasks, predicts continuous outcomes, or is a generative model (Section 6.1.2); whether a distance metric is available in the model inputs (Section 6.1.4); whether the model outputs binary or regressive values (Section 6.1.6); and whether base rates are equal (Section 6.1.8). Additionally, the fairness selection criteria also consider whether the data is biased (Section 6.1.3). Each of the fairness selection criteria are identified by a node in Figure 5.

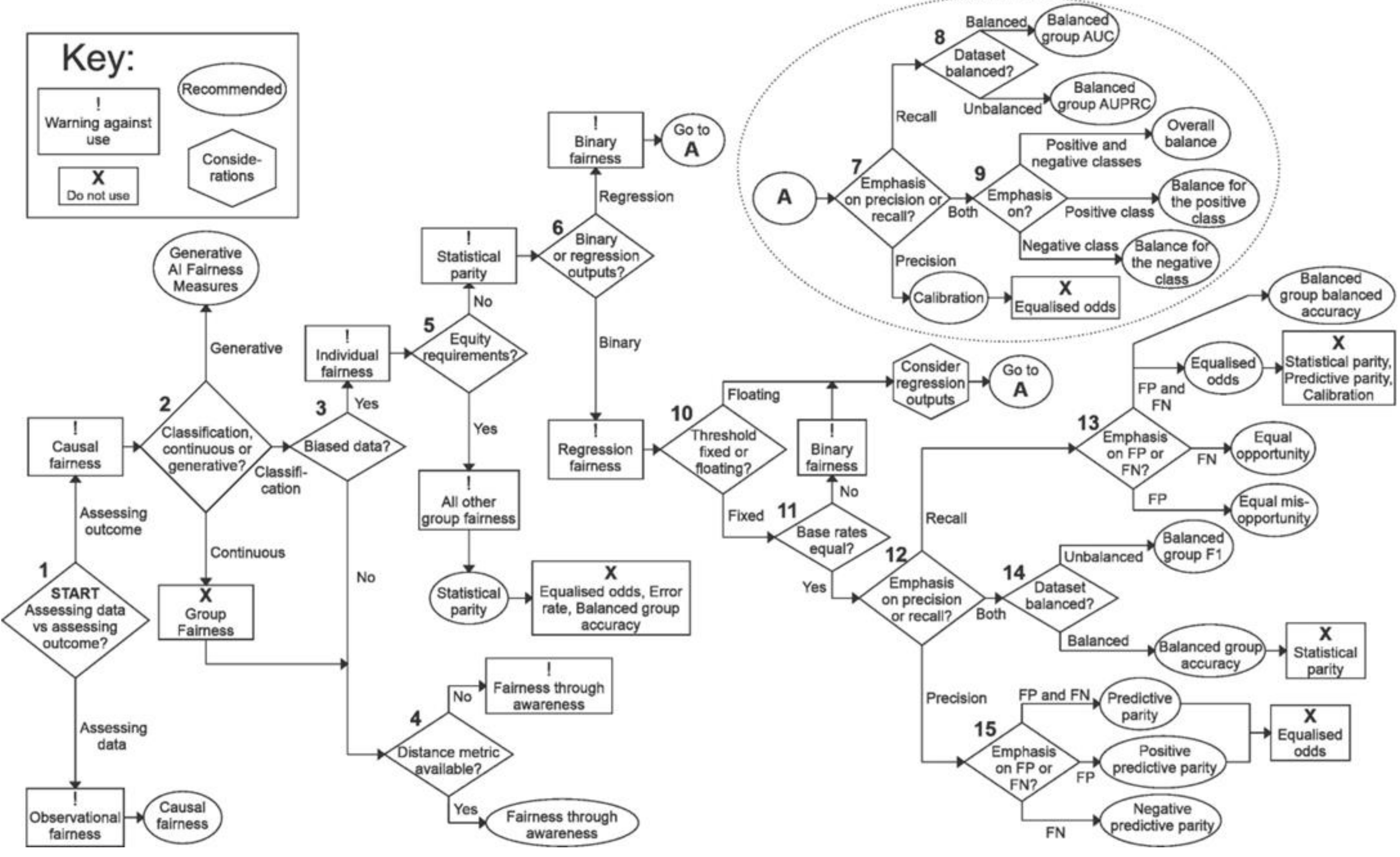

Figure 5: A flowchart for selecting context-appropriate observational fairness measures.

*6.1.1Assessing Data vs Assessing Outcome (Node 1)*

Tang et al. (2023) proposed that unbiased models achieve fairness requirements in data generation, predicted outcome, and induced impact. They argued that fairness in data generation be measured using causal definitions. Assessment of fairness in data generation requires consideration of the relationship between measured variables and the sensitive attribute only, avoiding any consideration of model formation. This is achievable by causal fairness measures as they account for the relationships between variables. They further proposed that fairness with respect to predicted outcome be measured using observational definitions, under the assumption that biases in the data are understood. This is because model predictions ($\hat{Y}$) are based on actual outcomes ($Y$), requiring $Y$ to be unbiased themselves. Observational fairness measures the relationship between $\hat{Y}$ and $Y$ directly [106].

*6.1.2Continuous Prediction, Classification, or Generative Modeling (Node 2)*

The type of ML model has important implications on the context-appropriateness of certain fairness definitions. Modern supervised ML has three broad categories: classification; continuous prediction; and generative modeling. When predicting on a continuous domain (e.g., predicting house prices), fairness through awareness is the only measure that can handle this data type as all other measures require group comparisons. Group fairness measures are more suited for classification tasks because binary outcomes are easily comparable between groups. Generative AI assesses fairness differently to common observational and causal measures. These measures primarily evaluate outputs based on specific prompts [163]. Assessment of fairness in generative AI is beyond the scope of the current paper and not considered here.

*6.1.3Biased Data (Node 3)*

Section 3 highlighted the complex ways bias can affect an AI model, with one of the main sources being the dataset itself. For AI models predicting binary outcomes with suspected bias in the dataset, individual fairness measures should be avoided. Individual fairness measures assess the similarity of predictions between similar cases, ignoring the relationship between $\hat{Y}$ and $Y$ (i.e., focusing only on whether overfitting or purposeful manipulation of $\hat{Y}$ has occurred). For instance, if historical biases are present, two individuals with the same input data ($X$) might have different $Y$ due to prejudices. Individual fairness measures would fail to detect this disparity because they only compare $\hat{Y}$ to $X$. In contrast, group measures account for the relationship between $\hat{Y}$ and $Y$, making them more appropriate in contexts where biased data is suspected. However, if there are no suspected biases in a dataset, individual measures of fairness can be effectively used.

*6.1.4Distance Metric Available (Node 4)*

When individual fairness measures are required, fairness through awareness is only possible when a distance metric between datasets can be calculated [34]. This is required to identify similar individuals for the calculation of (26).

*6.1.5Equity Requirements (Node 5)*

The US Equal Employment Opportunity Commission (EEOC) guidelines state that differences in probability of acceptance between two groups must not differ by more than 20% [164]. In hiring practices, equity requirements are commonplace around the world [165]. In any case when equity is desired, this is best measured by statistical parity notions [34]. However, when equity requirements are not mandated, statistical parity should be used with care as it does not account for the relationship between $\hat{Y}$ and $Y$.

*6.1.6Classification Models Output Binary or Regressive Values (Node 6)*

When outputs are continuous in a classification setting (regression), regressive measures of fairness are most appropriate [65]. Thresholding, required for binary fairness measures, can overlook the strength of predictions captured by continuous outputs. Additionally, thresholds can be manipulated to favor or reject different binary fairness notions depending on the modeler's biases. Larsen et al. [9] provides an example where binary fairness measures were incorrectly applied in a regressive scenario, highlighting the need for inclusion of this criteria in Figure 5. Conversely, for models that output binary results, only binary measures can be mathematically implemented.

*6.1.7Threshold Fixed or Floating (Node 10)*

Floating thresholds are used in cases where outcome requirements vary over time. For example, interest rates may change over time, influencing an individual's eligibility for a mortgage. In such cases, classification via regression should be considered, as fairness measures on the continuous domain remain invariant to temporal changes in thresholds. Conversely, when thresholds are fixed, all binary fairness measures become applicable, as they enable accurate comparison of classification outputs in relation to group membership [34].

*6.1.8Base Rates Equal (Node 11)*

Section 5.2 explained that when there is a significant disparity in base rates, and there is no apparent causal impact leading to this difference, binary fairness measures are inappropriate [18], [60], [108], [135]. In such cases, it is recommended to use regressive outputs. This is because fairness measures in regressive settings ignore the classification disparities caused by unequal base rates. However, when base rates are equal, selecting appropriate fairness measures requires additional contextual considerations.

*6.1.9Emphasis on Precision or Recall (Nodes 7/12)*

An emphasis on precision requires that positive predictions have high accuracy, whereas an emphasis on recall requires that positive outcomes are predicted with high accuracy. For example, considering death sentences has an emphasis on precision because it is crucial to ensure the offender is guilty, while predicting loan repayments emphasizes recall because incorrect predictions can be highly costly. When precision is prioritized, calibration and predictive parity are preferred because they incorporate the positive predictive value. When recall is prioritized, balanced group $AUC$, balanced group balanced accuracy, and equalized odds measures are preferred because they incorporate the true positive rate. When precision and recall are equally emphasized, balance of the positive and negative class, balanced group accuracy, and balanced group $F1$ can be used as they incorporate both the positive predictive value and the true positive rate [34].

*6.1.10 Emphasis on FP or FN (Nodes 13/15)*

Different scenarios can accept different $FP$ and $FN$ rates. For example, social media content filtering necessitates fewer $FP$'s because a $FP$ means that acceptable content is removed, upsetting the userbase. In contrast, diagnosis of potentially fatal illness necessitates fewer $FN$'s as a $FN$ will ultimately lead to untreated illness and mortality risk. When there is an emphasis on reducing $FP$'s, equal mis-opportunity and positive predictive parity are recommended because they emphasis fewer $FP$'s. When there is an emphasis on reducing $FN$'s, equal opportunity and negative predictive parity are recommended because they emphasis fewer $FN$'s. When fewer $FP$'s and $FN$'s are equally emphasized, balanced group balanced accuracy, equalized odds and predictive parity are recommended because they prefer fewer $FP$'s and $FN$'s [34].

#### *6.1.11 Emphasis on the Positive Class or Negative Class (Node 9)*

Other cases emphasize the rate, or size of the positive or negative class. Models that predict areas of high-risk criminal activity require predictions for the positive class to be fair because if these models disproportionately flagged areas with specific demographic groups, this could lead to the unfair targeting of certain demographics. Whereas school admission algorithms require predictions for the negative class to be fair to ensure that these models are not disproportionately rejecting certain demographics. When there is an emphasis on achieving fairness for the positive class, balance for the positive class is recommended. When there is an emphasis on achieving fairness for the negative class, balance for the negative class is recommended. When both balance for the positive and negative class are equally emphasized, overall balance is recommended [34].

#### *6.1.12 Balanced Dataset (Nodes 8/14)*

A balanced dataset has an equal number of positive and negative outcomes. Area under the precision-recall curve ($AUPRC$) and $F1$-score are best suited for unbalanced datasets. When datasets are balanced, it is recommended to use balanced group $AUC$ or balanced group balanced accuracy. When datasets are unbalanced, it is recommended to use balanced group $AUPRC$ or balanced group $F1$.

### 6.2 Using the Observational Fairness Selection Method

The creation of Figure 5 encapsulates the reviewed material on context-based fairness. The framework outlined in Figure 5 is designed to recommend appropriate observational fairness measures based on the specific context, while discouraging the use of less suitable measures. (Context appropriate selection of causal fairness measures are beyond the scope of this paper; for more information on this topic, refer to [33], [101], [166].)

Anahideh et al. (2022) proposed a framework that estimates the correlation among fairness notions for a given context using a Monte Carlo approach. These correlations were used to identify a set of diverse and distinct fairness metrics that best represent the given context [167], creating a subset of metrics that approximately concur and thus provide a "workable balance in the reduction of unfairness" [167]. The idea that fairness metrics may be correlated in a specific context is supported by Friedler et al. (2019), who also demonstrated that fairness metrics are correlated with each other [147]. However, this approach could fall into the 'circular argument' fallacy, where two correlated metrics could be equally inappropriate. Moreover, the implementation of certain fairness metrics in real-world applications is ultimately justified by a specific philosophical framework. Therefore, selecting fairness notions based on their proximity may not always be appropriate.

#### *6.2.1Limitations of the Observational Fairness Selection Method*

Figure 5 suggests context-appropriate fairness notions for measuring fairness of predicted outcome only (Section 6.1.1). These observational measures assess the relationship between model predictions, generated from the available dataset, and actual outcomes only. However, the available dataset may contain unwanted biases. This issue cannot necessarily be captured by the observational framework of Figure 5 and requires consideration of causal fairness. Causal fairness notions have been proposed for assessing fairness in data generation [106]. Tang et al. (2023) asserted that a fair model requires fairness in data generation, predicted outcome and induced impact [106]. Since the recommendations of Figure 5 did not consider fairness in data generation, Figure 5 could be augmented by the concepts provided by Makhlouf et al. (2022) [33] to select an appropriate causal measure of fairness. Fairness with respect to induced impact could then be observed by examining how users interact with the ML outputs, and the consequences of the model prediction.

#### *6.2.2Examples of Use*

Three examples are used to show how Figure 5 can be used to select an appropriate fairness measure.

##### *6.2.2.1 Prisoner recidivism*

Let us assume we are assessing fairness in predicted outcome (node 1). Classification models are used for predicting prisoner recidivism (node 2). Due to historically disadvantaged groups being arrested at higher rates [56], biases in the dataset would likely be present (node 3). Equity is not required for recidivism models (node 5). Let us also assume that binary outputs are being used (node 6) and that the threshold for recidivism is fixed (node 10). However, differing recidivism rates between races [39], [60], [62] would result in unequal base rates (node 11). Because of this, Figure 5 recommends the use of regressive outputs opposed to binary outputs. Therefore, instead of predicting whether an offender would reoffend, the model would now predict the likelihood of reoffending. Recidivism prediction models have an emphasis on recall (node 7) because incorrect classifications have high societal cost due to their use by parole boards. Assuming a balanced dataset (node 8) (i.e., equal number of reoffenders and non-reoffenders), balanced group $AUC$ would be recommended for assessing fairness.

##### *6.2.2.2 CV evaluation*

Assume we are assessing fairness in predicted outcome (node 1). CV evaluation models are classification models (node 2), with gender being the sensitive feature for fairness. Due to preconceptions, men are more likely to be hired than women with similar backgrounds [168], meaning biases may be present (node 3). Equity is required in CV evaluation models [26], [164] in most jurisdictions. Therefore, statistical parity would be recommended for assessing fairness.

##### *6.2.2.3 Spam filtering of political emails*

Let us again assume we are assessing fairness in predicted outcome (node 1). Spam filtering models are classification models (node 2). These models have been shown to be politically biased [169], meaning data biases may be present (node 3). Equity is not required by spam filtering models (node 5), and these models are binary (node 6) with fixed thresholds (node 10). Let us assume base rates are equal between political groups (node 11) (i.e., there is an equal rate of spam emailing from all political groups). There is an emphasis on both precision and recall (node 12) because spam filtering requires a high accuracy in both positive predictions, and for positive outcomes. Assuming datasets are balanced (node 14)—there is an equal number of spam emails to non-spam emails—balanced group accuracy would be recommended for assessing fairness.

### 6.3 Caveats of the Observational Fairness Selection Method

It is worth noting some caveats about Figure 5. Selbst et al. (2019) identified ‘portability’ as one of the failure modes fair ML researchers encounter. Researchers fall into a portability trap by failing to “understand how repurposing algorithmic solutions designed for one social context may be misleading, inaccurate, or otherwise do harm when applied to a different context” [69]. In effect, the prized goal of achieving a one-size-fits-all solution, highly valued in ML communities, often does more harm than good in fairness research due to the inability to apply a common measure of fairness across the broad range of ML problems. Whereas Figure 5 aims to mitigate the portability trap by using context to select an appropriate fairness measure, it may be too prescriptive in certain cases and ultimately ignorant of specific cases being considered. All fairness metrics must be carefully considered within the appropriate philosophical and cultural framework in which they are applied [22], [35]. Nonetheless, this potential for ambiguity should not inhibit the implementation of notions of fairness in ML. Ultimately, there is a moral, and often legal or commercial, impetus to introduce such notions when appropriate. Nor should the potential for ambiguity inhibit the generation of guidelines for selecting appropriate fairness metrics for specific contexts.

Fairness has a basis in philosophy, and thus definitions of fairness may change across cultures. Hence, the criteria introduced in Figure 5 should be carefully considered against the cultural framework where the metric is to be applied. This paper considered published literature predominantly from Western countries. Thus, care must be taken to avoid the portability trap by implementing certain fairness notions across cultures without careful input from members of those cultures.

## 7 CONCLUSION

Concerns with fairness in ML have grown in both academia and governance. New regulations are being proposed globally to ensure that AI is both unbiased and fair [23], [25]. This demand has motivated research into defining context-appropriate fairness notions [17], [22], [31], [32]. Both academia and standardization have contributed to this, defining bias and fairness in AI independently. Despite this, the reviewed material shows that biases defined in the ISO/IEC TR 24027:2022 [28] are similar to definitions in academic literature. Our review found that bias interacts with AI systems in complex ways. Assessment of bias thus requires knowledge of the context surrounding data collection, development and interaction within AI systems (Figure 3).

A systematic review of recent papers identified equal opportunity, statistical parity, and equalized odds as the most common fairness measures (Table 1). Causal measures, which are now seen by some as essential to address fairness [33], [97], [98], were not identified in any of the studies reviewed. This may be due to the difficulty of employing causal measures of fairness [21]. While many fairness measures can be applied in specific scenarios, each measure has trade-offs with model performance that are context dependent [16].

The contextual appropriateness of certain fairness measures in given scenarios is further restricted by unequal base rates [18] and fairness notion incompatibilities [30]. Under these specific scenarios, certain fairness measures are inappropriate. Therefore, a flowchart was developed to guide the selection of contextually appropriate observational fairness measures (Figure 5) to identify fairness in predicted outcome. When Figure 5 is used alongside causal notions for assessing fairness in data generation, a fair ML approach is achievable [106].

Twelve criteria were used to formulate Figure 5. This included consideration of the model assessment criteria, design and data. The flowchart was used to determine contextually appropriate observational fairness notions for three ML applications: prisoner recidivism, CV evaluation, and political spam filtering. In these applications, Figure 5 recommends the use of balanced group $AUC$, statistical parity, and balanced group accuracy for assessing fairness respectively.

Careful implementation of Figure 5 can be used to assist lawmakers, modelers, and researchers in identifying context-appropriate fairness notions for measuring fairness of predicted outcome. Future research should consider the implementation of fairness beyond a Euro-centric lens to ensure such measures are implemented in accordance with the local culture and legal frameworks. Additionally, development of a method for context-appropriate selection of fairness measures in generative AI is crucial, given the rapid expansion of this field and its significant impact on daily life.